\documentclass[11pt,a4paper]{article}
\usepackage[T1]{fontenc}
\usepackage{authblk}
\usepackage[letterpaper,bmargin=1in,tmargin=1in,lmargin=1in,rmargin=1in]{geometry}
\usepackage{listings}
\usepackage{color}
\usepackage{hyperref}
\usepackage[table]{xcolor}
\usepackage{graphicx}
\usepackage{multirow}
\usepackage{amsmath,amssymb,amsbsy}

\usepackage{amsfonts}

\title{Regularization and feature selection for large dimensional data}
\begin{document}

\author[1]{Nand Sharma\thanks{sharma@math.colostate.edu}}
\author[2]{Prathamesh Verlekar}
\author[2]{Rehab Ashary}
\author[2]{Sui Zhiquan}
\affil[1]{Department of Mathematics, Colorado State University, CO, USA}
\affil[2]{Department of Computer Science, Colorado State University, CO, USA}

\date{}


\maketitle

\begin{abstract}
Feature selection has evolved to be an important step in several machine learning paradigms. In domains like bio-informatics and text classification which involve data of high dimensions, feature selection can help in drastically reducing the feature space. In cases where it is difficult or infeasible to obtain sufficient number of training examples, feature selection helps overcome the curse of dimensionality which in turn helps improve performance of the classification algorithm. The focus of our research here are five embedded feature selection methods which use either the ridge regression, or Lasso regression, or a combination of the two in the regularization part of the optimization function. We evaluate five chosen methods on five large dimensional datasets and compare them on the parameters of sparsity and correlation in the datasets and their execution times.
\end{abstract}

\section{Introduction}

Guyon and Elisseeff \cite{feat} classify feature selection methods into three broad categories: \textsl{Filter}, \textsl{Wrapper} and \textsl{Embedded} methods. In Filter methods, features are scored typically by some measure of correlation with the labels. Examples of this method are Pearson correlation, Fisher scores, Golub coefficient, and Uncorrelated Shrunken Centroid (SC). Wrapper methods use a predictive model in order to score feature subsets, and each subset is used to train a model. Testing process here is performed on the hold-out test set and counting the number of mistakes that occur during the testing process gives the score for the subset. Examples of this method are Sequential forward search (SFS), Decision tree + SVM, Rough Set + SVM \cite{Liu2006,comparison2008} etc. Embedded methods use properties of the classifier and perform feature selection within the training process. Recursive Feature Elimination (RFE) using L2 norm SVM, L1 Norm SVM, and Local Learning (LL) are some examples of the embedded methods.\\

In recent times, the role of regularization in feature selection has become very prominent. L2 (ridge) regularizer has been widely used because of its simplicity and proven capability at two class classification. But when the data is high dimensional and there are many redundant noise features, L1 (Lasso) regularizer has proven to work better. L1 regularization is very effective in reducing dimensionality, thus encouraging sparsity of models that can be more easily interpreted. The focus of our research are five embedded feature selection methods to study the affect of regularization on feature selection. These methods either use L2 norm, or L1 norm  or a combination of the two for regularization.

Support vector machines (SVM) have been extensively used as a state of the art classifiers and are often combined with various feature selection methods to obtain better classification with optimal feature subset. \cite{Chapelle08} focuses on comparing only embedded methods namely RFE and L1 SVM. \cite{comparison2008,Liu2006,Chen05combiningsvms} compare filter and wrapper feature selection methods by combining SVM with various feature selection strategies but do not include comparison with any embedded methods.  In our study, we compare five embedded feature selection methods, and also compare these methods with two filter methods - SC and Golub.

Another novelty of this study is that we apply these algorithms on several datasets from different domains to offset too much dataset dependence of results. The datasets used in our study vary from cancer patients data to text classification to optical character recognition (OCR). \cite{comparison2008} uses two datasets with maximum of 24 features in them while \cite{guyon2004result} is an extensive review paper dealing with many datasets but the maximum any dataset contains is 30000 features. We have used a gene microarray dataset, Dorothea, from the NIPS 2003 feature selection challenge which has 100000 features in addition to four other large dimensional datasets of sparse and dense nature.   

Feature selection can be a very cost-intensive process. Getting the optimal subset of features also depends on the execution time when the dataset has very high dimensionality. So, we have compared algorithms in terms of execution time also.

\section{Regularization and Feature selection methods}
Regularization plays an important role in Feature selection methods that are embedded. To motivate the concept of regularization, we consider the usual linear Least Square regression \cite{trevor2001elements} which is one of the popular methods for classification: given training data $\lbrace x_{1}, x_{2}, \dots ,x_{n}\rbrace; x_i \in \textbf{R}^{d}$ and the associated class labels $\lbrace y_{1}, y_{2}, \dots ,y_{n}\rbrace; y_i \in \textbf{R}$, the traditional least squares regression(LS) solves the following optimization problem to obtain the weight vector ${w}\in \textbf{R}^{d}$ and the bias $b\in \textbf{R}$

\begin{equation}\label{eq1}
\min_{w,b} \sum \limits_{i=1}^{n} ||{w^T x_i + b - y_i}||_{2}
\end{equation}

When the dataset has  large number of features compared to the number of observations, which is the case we are interested in, that is $d \textgreater \textgreater n$,  Eq. \ref{eq1} produces a poor estimation due to the high variance of the estimated weight coefficients. Moreover, there is the problem of overfitting because of the large potential of modeling the noise. These lead to poor performance of LS in both prediction and interpretation.  

Penalization techniques have been proposed to improve LS. For example, ridge regression (Hoerl and Kennard, 1988) minimizes the residual sum of squares subject to a bound on the L2-norm of the coefficients. As a continuous shrinkage method, ridge regression achieves its better prediction performance through a bias-variance trade-off \cite{bishop}. Thus the problem becomes 

$$
  \min_{w,b} \sum \limits_{i=1}^{n}||{w^T x_i + b - y_i}||_{2} + \lambda||{w}||_{2}
$$

The above function being minimized can be seen as a sum of two parts : the \textit{loss function} and the \textit{regularizer}. The general binary classification problem on similar lines can be written as 

$$
\min_{w,b} F(w,b) = \min_{w,b} L(w,b) + \lambda\ R(w,b)
$$
where $L(w,b)$ is the loss function and $R(w,b)$ is the regularizer, and $\lambda$ is the tuning parameter that controls the trade-off between loss and regularization. Examples of popular loss functions are the Hinge loss, Log loss and exponential loss, while examples of popular regularizers are the general class of \textit{$l_p$} norm.

In this paper we analyze regularization by studying five embedded methods that use different kinds of regularizers and/or loss functions. We also compare these methods with two filter methods for reference. A brief description of each of the methods follows.

\paragraph{ L{\scriptsize 2} Norm SVM}
We start our study of the methods chosen with the standard \textit{$l_2$} norm Support Vector Machine (SVM). We use SVM as the common classifier for all the feature selection methods and also as a feature selection method in combination with the method of Recursive Feature Elimination (RFE).
 
For a binary classification problem, the SVM finds a separating hyperplane with maximal margin between the two classes. The standard \textit{$l_2$} norm SVM uses the Hinge loss function and a \textit{$l_2$} norm regularization, so takes the following form

$$
\min_{w,b} C\sum_{i=1}^n \xi(w,b;x_i, y_i) + \lambda ||{w}||_{2}  
\label{eq4}
$$

where $\xi(w,b;x_i, y_i)$ is the loss function and $C\geq 0$ is a penalty on the training error. The loss function used here is the hinge loss defined by
\begin{eqnarray*}
 \xi(w,b;x_i, y_i) = max(1-y_i(w^T\phi(x_i)),0)
\end{eqnarray*} 
where $\phi$ is the basis function used to map the data to higher dimensional space. Then the kernel trick helps classification in this higher dimensional space without additional computation cost.
For this a kernel function $K(x_i,x_j)= \phi(x_i)^T\phi(x_j)$ is used. For our classification , we use the linear SVM where $\phi(x)= x$, and the Gaussian radial basis function (RBF) for which 
$K(x_i,x_j)= exp(-\gamma||{x_i-x_j}||_{2}) , \gamma>0$.

\paragraph{ L{\scriptsize 1} Norm SVM}
The second method we consider is the \textit{$l_1$}-SVM (L1) (\cite{kujala}). This method replaces the standard \textit{$l_2$}-norm penalty with the \textit{$l_1$}-norm penalty, and the 1-norm SVM obtained is 

$$
\min_{w,b} C\sum_{i=1}^n \xi(w,b;x_i, y_i) + \lambda ||{w}||_{1}  
$$

The \textit{$l_1$}-norm SVM has some advantages over the standard 2-norm SVM, especially when the redundant noise features are considered. A noticeable fact is that the 1-norm penalty is not differentiable at zero\cite{zhu}. This important singularity property ensures that the 1-norm SVM is able to delete many noise features by estimating their coefficients by zero. One drawback of using \textit{$l_1$}-norm SVM is that the number of selected features is bounded by the number of samples. It has also been observed that for sparse data, \textit{$l_1$}-SVM does not perform very well. This might be due the strong correlation between some features in the data which \textit{$l_1$}-SVM fails to pick. 

\paragraph{Local Learning Based Feature Selection}
Another method that uses \textit{$l_1$} penalty is a `local learning' based method (LL)\cite{Sun_locallearning}. LL  is based on the concept of decomposing a given nonlinear problem into a set of locally linear problems. This method also maximizes a margin like the SVMs but the margin here is defined differently. It is defined for each training sample in terms of its two nearest neighbors, one from the same class (called nearest hit or NH), and the other from the different class (called nearest miss or NM). Each feature is then scaled to obtain a
weighted feature space, giving the margin of $x_{n}$, computed with respect to w as:

\[
\rho_{n}(w)=\textit{d}\left(x_{n},NM\left(x_{n}\right)|w \right) -\textit{d} \left(x_{n},NH\left(x_{n}|w\right) \right) = w^{T}z_{n}
\]
 
where $z_{n} = |x_{n} - NM\left(x_{n}\right)| - |x_{n} - NM\left(x_{n}\right)|$

The nearest neighbors for this purpose are obtained using a probabilistic model, where the nearest neighbors of a given sample are treated as hidden variables and are estimated using expectation-maximization over the whole training set.

This method uses the logistic regression formulation, and hence a log loss function for maximizing the margin, and \textit{$l_1$} regularization, which leads to the following optimization problem:

\[
\min_{w}\sum\limits_{n=1}^N log\left( 1 + \exp\left(-w^{T}{z}_{n}\right) \right) + \lambda \parallel w \parallel_{1}
\]

\paragraph{Elastic net}

The elastic net (EN) penalty was proposed in \cite{Zou05regularizationand} to overcome the limitations of the lasso penalty, namely that the lasso selection method ($l_1$ norm) is restricted by the number of observations. The elastic net penalty has been shown to produce a sparse model and to encourage a grouping effect (that is, the correlated features as a whole will be in or out of the model). The naive elastic net regularization is a convex combination of $l_1$ and $l_2$ norm and is defined as:
\[
 \lambda_{1} ||w||_1+ \frac{\lambda_{2}}{2} ||w||^2_2 , 
\]

where the $l_1$ norm encourages a sparse model and the $l_2$ norm encourages continuous shrinkage by shifting the coefficients toward zero (but never exactly zero). In other words, the coefficients are shrunk towards each other. Therefore, the $l_2$ norm forces the learning algorithm to assign similar coefficients to correlated features. As a result, it shows the property of grouping effect to ensure that correlated features are either in or out of the model as a group. The elastic net therefore displays a nice property whereby it retains the lasso's main property (i.e. sparsity) and resolves its difficulties where the number of selected features is not bounded by the number of observations.

\paragraph{ L{\scriptsize 2,1} Norm Minimization based Feature Selection}

This feature selection method (L21) was proposed to be used for both binary and multi-task learning \cite{method} with the consideration for the correlation between the features\cite{use}.

The \textit{$l_{2,1}$} norm of a matrix $\textbf{M} = (m_{ij})$ was first introduced in [5] as rotational invariant \textit{$l_2$} norm and also used for multi-task learning [1, 18] and tensor factorization [10].It is defined as 
$$
||M||_{2,1}= \sum_{i=1}^n ||m^i||_{2}
$$
This feature selection method is based on \textit{$l_{2,1}$} norms for both loss function and regularization term. The loss function that the authors call 'robust loss function' is built on the least squares regression method, with the difference that the residual error is not squared which results in the outliers having less importance. 

Designed basically as a general multi-class classification, for class labels $\lbrace y_{1}, y_{2}, \ldots ,y_{n}\rbrace\in \textbf{R}^{c}$, and the weight matrix ${W}\in \textbf{R}^{d\times c}$, and $b \in \textbf{R}^{c}$, the optimization problem is given by  

$$
\min_{w,b} \sum_{i=1}^n ||{W^T x_i +b - y_i}||_{2,1} + \lambda||{W}||_{2,1}
$$

\paragraph{Reference Filter methods}
The Uncorrelated shrunken centroid (SC) \cite{usc} and Golub feature scoring method (Golub) \cite{Golub99molecularclassification} are used as reference filter methods for our study.

\section{Experimental setup}\label{l3}

\subsection{Data Sets}
We used sparse and non-sparse high dimensional data sets for comparison. These data sets are taken from different domains, and hence the feature selection performance can be evaluated on various applications. Four data sets are taken from NIPS feature selection challenge \cite{guyon2004result,guyon2003design}. The fifth data set is Arabidopsis. The Arabidopsis data addresses a prediction problem in the area of gene splicing:  when mRNA in the cell is made, it first undergoes a process called splicing in which parts of a gene called introns get removed and the rest (called exons) gets spliced back together. A given gene can be spliced in multiple ways, a process called alternative splicing (\cite{arb1}).  In some cases a given intron might not undergo splicing, in which case, it remains in the final mRNA.  This is known as intron retention, and is the primary form of alternative splicing in plants.  We assembled a dataset consisting of XXX introns that are annotated as retained, and YYY introns that do not exhibit this phenomenon.  The DNA sequence of each intron was represented by its composition of substrings of length 6 (also known as the spectrum kernel), where we also included features that represent substrings of length 6 present in the intron's flanking exons.  As intron retention is a process regulated by proteins recruit or suppress splicing, we expect this representation to capture signals that drive the underlying biological process (\cite{arb1}).  Feature selection would then provide us clues on patterns that are associated with this process.

Dexter, Dorothea, and Arabidopsis are sparse datasets while Arcene and Gisette are dense in nature. Arcene contains mass-spectrometric of three cancer data set, namely one ovarian and two prostate data. Arcene dataset is used widely to distinguish cancer versus normal patterns. This is a two-class classification problem with continuous input variables. Dexter is a text classification problem in a bag-of-word representation where the real features represent the commonness of word in the texts (corporate acquisition and other topic texts). This is also a two-class classification problem with sparse continuous input variables. Dorothea is a data set dealing with gene splicing and binding sites. The task of this data is to foresee which compound bind to a target site on Thrombin. Gisette is a handwritten digit recognition problem. The task of this data is to separate the digits 4 and 9 which becomes a two-class classification problem . Besides the actual features , these datasets contain a number of non-informative features (probes) that mislead the classifier. 

The probes in the Arcene dataset are constructed by permuted least informative features. The Zipf law is utilized to infer the probes features in the Dexter data. The probes in Dorothea are created by permuting the last 50000 of 100000 features ranked using Weston et al. standards \cite{guyon2003design}. The probes in the Gisette data represent permuted pair.

A further characteristic of these data sets is that the number of features is larger than the number of examples with the exception of the Gisette. Table \ref{tab1} illustrates the number of examples, the number of real features and probes, the percentage of sparsity and average absolute pairwise correlation.

\begin{table}[h!]
\centering
\begin{tabular}{|c p{1.5cm} p{1.5cm} p{1.5cm} p{1.5cm} p{1.5cm} p{1.5cm} p{1.6cm} |}
\hline 

Dataset & Training Examples & Validation Examples & Test \hspace{15 mm}Examples & Real \hspace{15 mm}Features & Probes & Sparsity & Correlation 

 \\ \cline{1-8} 
\multirow{1}{*}{Arcene} & 100 &   100& 700 & 7000  & 3000 & 50$\%$&0.1831
  \\
 \cline{1-8}
\multirow{1}{*}{Dexter}&300&   300 & 2000 & 9947   & 10053 &99.5$\%$& 0.0137
\\
 \cline{1-8}
\multirow{1}{*}{Dorothea}& 800 &   350  &800 & 50000  &  50000 &99$\%$ &  0.7882
\\
 \cline{1-8}
\multirow{1}{*}{Gisette}& 6000 &   1000  &6500 & 2500  &  2500 &87$\%$& 0.0222
\\
 \cline{1-8}

\multirow{1}{*}{Arabidopsis}& 5827 &   1166 & 4661 & 16390 & 0 &96.5$\%$& 0.0102
\\
 \cline{1-8}
 \hline
\end{tabular}
\caption{The Arcene dataset has 44 positive and 56 negative examples while the positive and negative examples are balanced distributed among the training, the validation, and the test sets of the Arabidopsis data. The Dorothea has 78 positive and 722 negative examples in the training and test sets and 34 positive and 316 negative examples in the validation set. The other datasets hold an equal number of positive and negative examples.}
\label{tab1}
\end{table}

\subsection{Experiments}
To test the efficiency of the algorithms for feature selection, we run these algorithms on the five datasets mentioned above to select features, and then classify the data using SVM as a common method for classification. 

The following are the steps followed for the embedded and wrapper methods- Elastic Net, Local Learning, L1 and L21 SVMs and SVM RFE.

\begin{enumerate}

\item We use the training and validation sets to optimize the parameters for each algorithm. This is done using classification accuracy (balanced success rate or BSR) of each algorithm on the validation set.

\item The validation and training sets are combined to detect the features using these optimized parameters for each algorithm.

\item The top ranked 1000 features are selected for model selection of SVM using 5-fold cross-validation on the datasets comprising both training and validation sets.

\item Classification of test data is performed using SVM with parameters chosen during the model selection step.

\end{enumerate}

For the filter methods - SC, and Golub, the following steps are followed. 

\begin{enumerate}
\item We use the combination of the validation and training sets to get the ranking of features based on the feature scores.

\item The top 1000 features were selected for model selection of SVM using 5 fold cross-validation.

\item The test data is then classified using the SVM with the model selected during the validation process.

\end{enumerate}

For model selection using SVM, which is a common step for both types of algorithms, we use all combinations of values of C with Gaussian and linear kernels. The C values used are 0.1, 1, 10 and 100. For the Gaussian kernel we used the kernel with $\gamma$ values of 0.005, 0.02, 0.5 and 2.0.

The balanced success rates for each of the algorithms was recorded and compared at various numbers of features selected. We also compared the execution time of the algorithms.

PyML\cite{py} was used for SVM classification. The linear SVM was found to perform the best, so the liblinear version of SVM was used for final classification. 

\section{Results and Discussion}\label{l4}
\begin{figure}[!htbp]
\begin{center}
\includegraphics[width=6in]{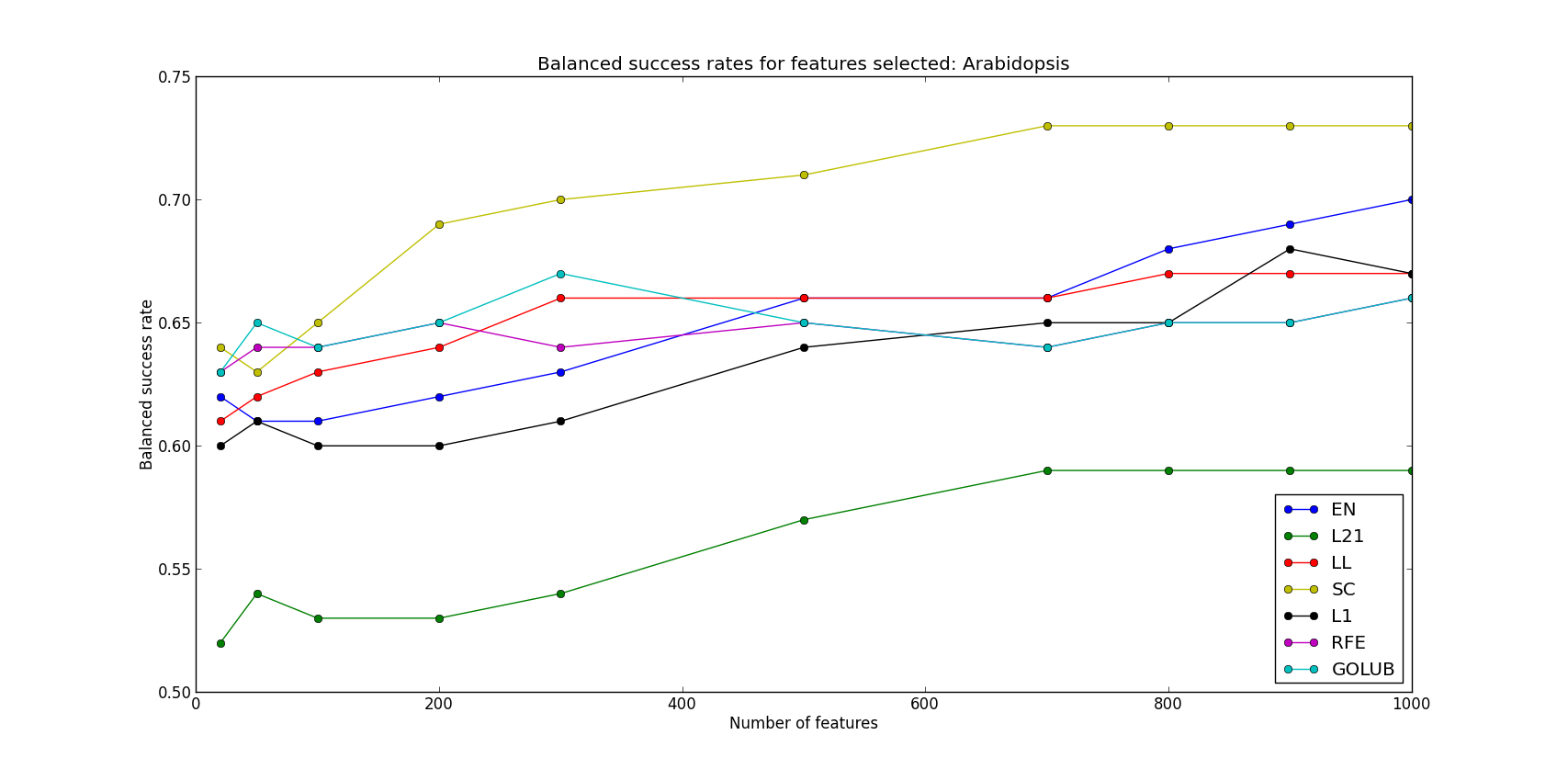}
\caption{BSR for top 1000 features(Arabidopsis)}
\label{BSR_arabidopsis}
\includegraphics[width=6in]{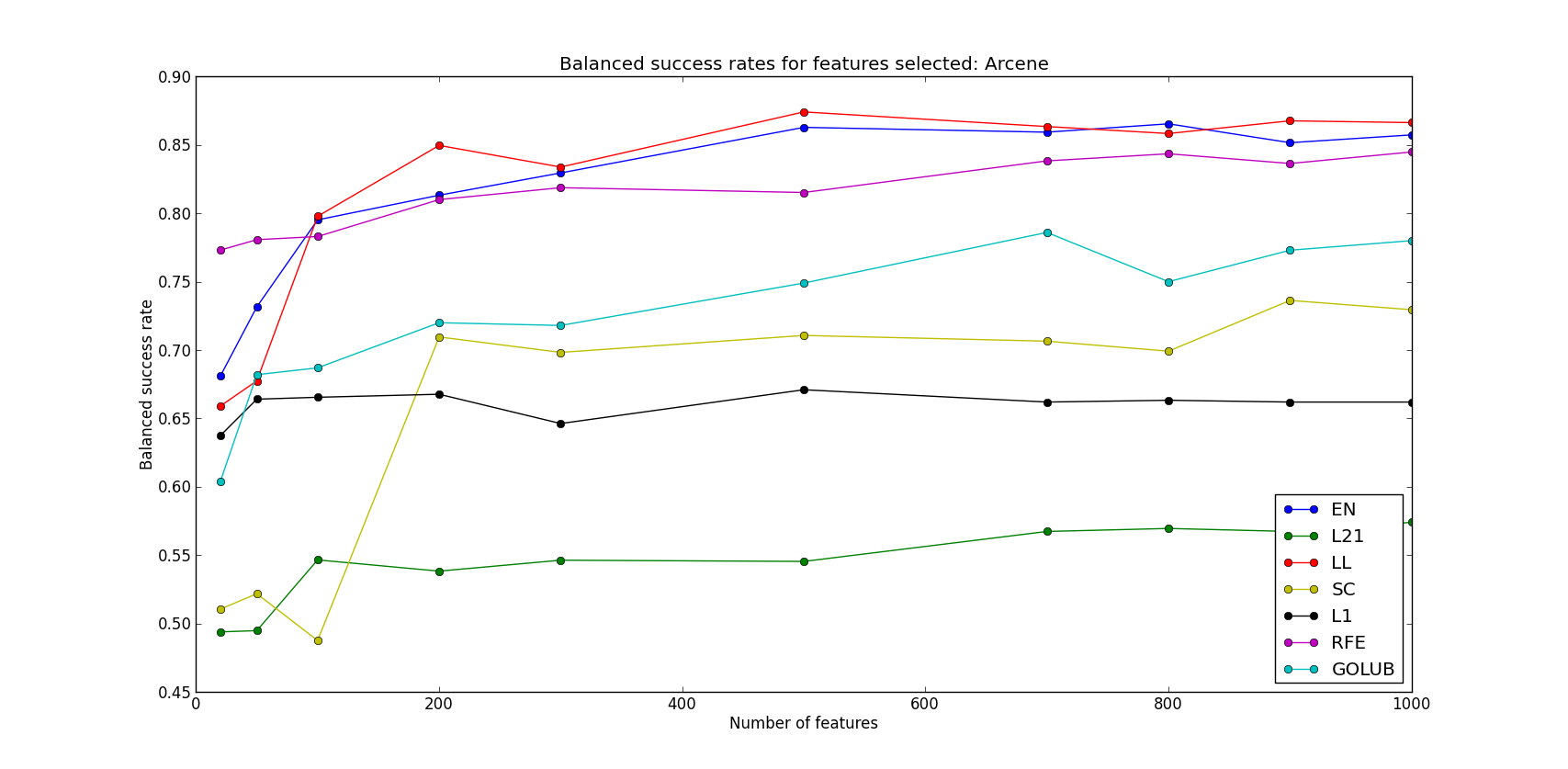}
\label{BSR_arcene}
\caption{BSR for top 1000 features(Arcene)}
\includegraphics[width=6in]{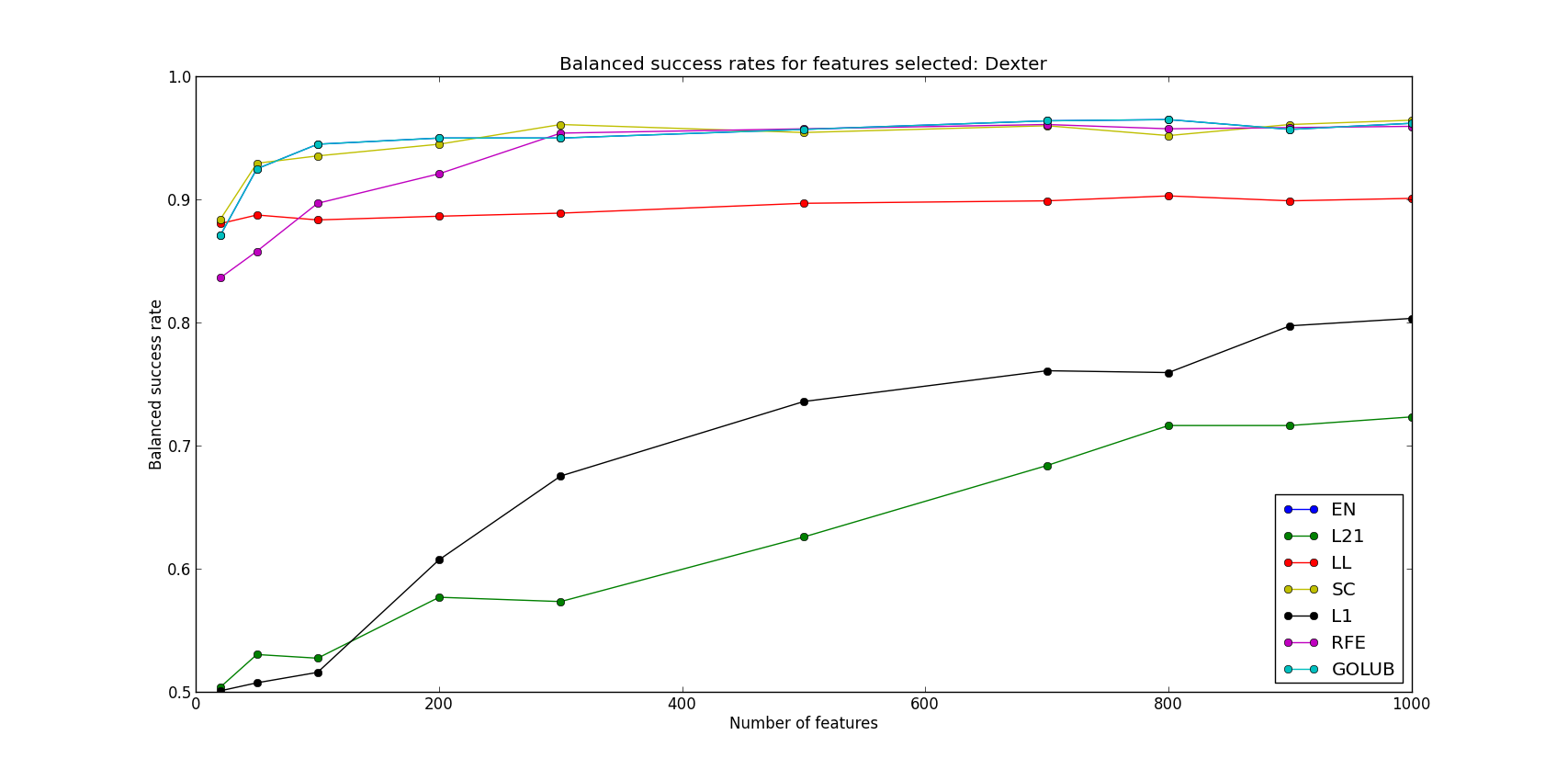}
\caption{BSR for top 1000 features(Dexter)}
\label{BSR_dexter}
\end{center}
\end{figure}

\begin{figure}[!htbp]
\includegraphics[width=6in]{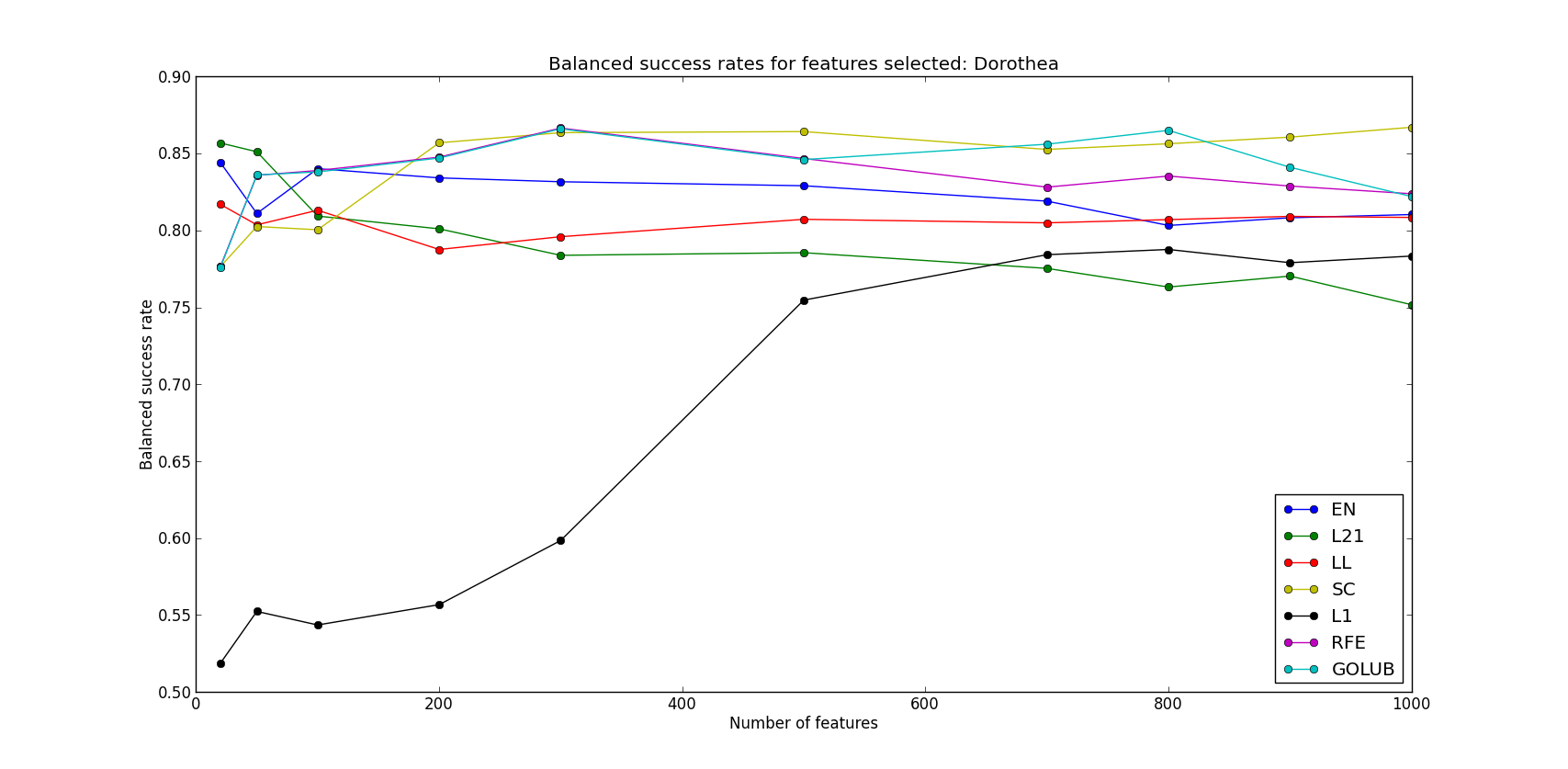}
\caption{BSR for top 1000 features(Dorothea)}
\label{BSR_dorothea}
\includegraphics[width=6in]{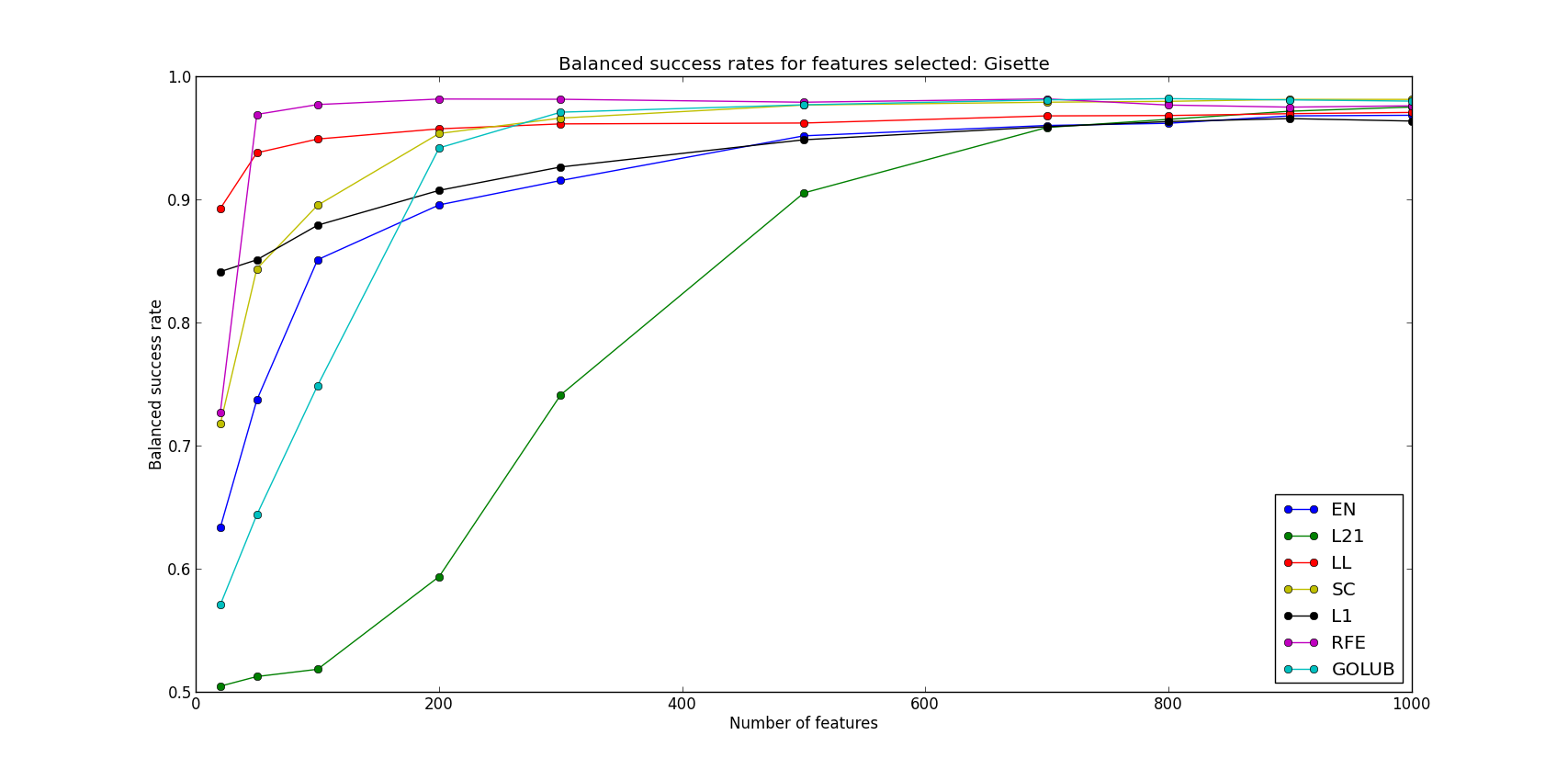}
\caption{BSR for top 1000 features(Gisette)}
\label{BSR_gisette}
\end{figure}

For our analysis, we look at BSR at 50 and 200 features in tables \ref{table: 50feat} and \ref{table: 200feat}, and plots of the BSR against the number of features selected for each of the datasets are as shown in figures \ref{BSR_arabidopsis}, \ref{BSR_arcene}, \ref{BSR_dexter}, \ref{BSR_dorothea}, and \ref{BSR_gisette}. For every algorithm, an increase in the magnitude of the BSR was observed for all the datasets when the number of features were increased from 50 to 200. As we observe the plots, we find that the relative performance of algorithms across datasets is more or less stable beyond 200 features. 


So, the initial analysis was performed by isolating datasets and studying their results at 200 features. For Arabidopsis, highest accuracy was achieved for SC method with 200 features selected. In general for Arabidopsis, the BSR increased for 5 of the 7 algorithms when the number of features were increased from 50 to 200. For Arcene, it was observed that LL method gave best BSR for 200 features selected. For Dexter, it was SC method which performed best for both 50 and 200 features, though EN beat SC by 0.005\% for 200 features. In case of Dorothea, it was observed that for 50 features, L21 gave the best BSR while for 200 features, it was SC which worked the best although the BSR values were around 0.85 for both the cases. Finally, for Gisette, RFE was the best performer for both the analysis i.e. involving 50 features and 200 features respectively. For high dimensional data and non sparse data, RFE gives best results.

\begin{table}[!htbp]
\caption{Balanced Success Rates for top 50 features(Percentage of probes retained in braces)}
\centering
\begin{tabular}{c c c c c c}
\hline\hline
Datasets $\longrightarrow$ & Arabidopsis & Arcene & Dexter & Dorothea & Gisette\\ [1ex] 
Algorithms $\downarrow$ \\
\hline 
L1 & 0.61 & 0.6641(38) & 0.5075(26) & 0.5550(52) & 0.8511(62) \\
LL & 0.62 & 0.6775(28) & 0.8875(46) & 0.8036(60) & 0.938(48) \\
EN & 0.61 & 0.7316(56) & 0.9255(0) & 0.8110(18) & 0.7372(0) \\
L21 & 0.54 & 0.4949(28) & 0.5305(8) & \textbf{0.8511}(40) & 0.5126(48) \\
RFE & 0.64 & \textbf{0.7807}(38) & 0.858(2) & 0.8358(0) & \textbf{0.9692}(52) \\
SC & 0.63 & 0.5219(32) & \textbf{0.9295}(2) & 0.8025(0) & 0.8438(58) \\
GOLUB & \textbf{0.65} & 0.682(34) & 0.925(0) & 0.836(0) & 0.644(50) \\[1ex]
\hline
Baseline & 0.6946 & 0.8756 & 0.9665 & 0.5 & 0.9775 \\
\hline
\end{tabular}
\label{table: 50feat}
\end{table}

\begin{table}[!htbp]
\caption{Balanced Success Rates for top 200 features(Percentage of probes retained in braces)}
\centering
\begin{tabular}{c c c c c c}
\hline\hline
Datasets $\longrightarrow$ & Arabidopsis & Arcene & Dexter & Dorothea & Gisette\\ [1ex] 
Algorithms $\downarrow$ \\

\hline 
L1 & 0.60 & 0.6671(43.5) & 0.6075(33) & 0.5374(53.5) & 0.9075(53) \\
LL & 0.64 & \textbf{0.8496} & 0.8865(52.5) & 0.7876(59.5) & 0.9575(52.5) \\
EN & 0.62 & 0.8132(52.5) & \textbf{0.950}(10) & 0.8341(52) & 0.8957(0) \\
L21 & 0.53 & 0.5384(34) & 0.577(13.5) & 0.801(76) & 0.5938(51.5) \\
RFE & 0.65 & 0.81(32) & 0.921(7.5) & 0.8476(36) & \textbf{0.9817}(49) \\
SC & \textbf{0.69} & 0.7096(28.5) & 0.945(8.5) & \textbf{0.8569}(0) & 0.9537(54.5) \\
GOLUB & 0.65 & 0.72(30.5) & 0.950(10) & 0.847(36) & 0.942(53.5) \\ [1ex]
\hline
Baseline & 0.6946 & 0.8756 & 0.9665 & 0.5 & 0.9775 \\
\hline
\end{tabular}
\label{table: 200feat}
\end{table}

The five datasets have different properties: Dexter, Dorothea, and Arabidopsis are sparse; Arcene and Gisette being non-sparse. Arcene is the only continuous valued data set. Gisette and Dorothea are the datasets with highest dimensions (100,000) used in these experiments. As seen in the graphs and tables above, for sparse datasets, SC is the best with the highest BSR except with Dorothea which has the highest number of features; L21 gave best result with less number of features while on the other hand, SC method gave the highest BSR with more number of features. These results can be justified by the fact that these datasets consist of many redundant features and the nature of SC method removes these redundant features. With the non-sparse data sets, either with continuous variables (Arcene) or high dimensional data set (Gisette), LL method gave the highest BSR over all the presented methods (if RFE and GOLUB are not considered as they are for reference in this work).

Methods EN and L21 both use a combination of L1 and L2 norms. But in EN method, both L1 and L2 norms are used in the regularization part alone while L21 method uses a combination of the two in both the regularization part and the loss function. As shown in the tables and figures, in most cases EN gave higher BSR than L21 algorithm indicating that using both L1 and L2 norms with regularization part only is more productive than using it with both loss function and regularization. This appears to be a strong point of EN.

Now, we look at the correlation in the data and see how that impacts the performance of our algorithms. As expected, L1 doesn't do well for datasets with high values of correlation. On Dorothea, which has the highest correlation, L1 in fact performs the worst. This is on the expected lines. Again, SC which is method that is based on the correlation does in fact give the best results as we would expect. But among the methods that we are comparing, EN works the best, followed by L21.

On Arcene, which has the next highest correlation in data, LL is the best performer, but EN is again quite close. Interestingly L21 is not so good. This might again be explained by the fact that using both L1 and L2 norms with regularization part only is more productive than using it with both loss function and regularization.

On datasets with low values of correlation, LL works the best in two out of the three datasets (Gisette, Dexter, Arabidopsis). L1 is the next best on two. This is again explained by the nature of L1 that it works best on less correlated datasets. Again though EN is the best performer in one of these 3 and the second best in one. This again seems to indicate the `versatility' of EN for both highly correlated and less correlated data.  

Now, we look at the performance of the algorithms with respect to sparsity of data recorded in table \ref{tab1}. Results show that for dense datasets, Local Learning, L1-SVM and EN gave the best results, while Elastic Net, Local Learning and SVM-RFE gave the best results for sparse datasets. 

Another performance measure that was used is the execution time that a given algorithm took to compute the list of relevant features. The table \ref{table: exectime} shows the results. In many very high-dimensional data, execution time can sometimes be of important consequences. The table shows that among the algorithms that we compare, EN stands out as exceptionally better than other algorithms. Next best is L1.

\begin{table}[!hbt]
\caption{Average execution time(in seconds)}
\centering
\begin{tabular}{c c c c c c}
\hline\hline
Datasets $\longrightarrow$ & Arabidopsis & Arcene & Dexter & Dorothea & Gisette\\ [1ex] 
Algorithms $\downarrow$ \\

\hline 
L1 & 100.32 & 158.322 & 381.141 & 29.699 & 1659.465 \\
LL & 8386.2 & 4.0401 & 20.5494 & 2735.7 & 2850.2 \\
EN & 5.333  & 0.774 & 0.954 & 9.842 & 10.217 \\
L21 & 2639.3 & 1358.23 & 5309.57 & 306.5 & 1065.9 \\
RFE & 8096.881 & 42.703 & 23.424 & 607.827 & 4017.653 \\
SC & 37112.07 & 291.81 & 598.37 & 235512.80 & 2743.86 \\
GOLUB & 45.997 & 6.148 & 0.944 & 12.157 & 112.399 \\ [1ex]
\hline
\end{tabular}
\label{table: exectime}
\end{table}

\section{Conclusion}
It's always hard to state definitive conclusions in any comparative study of algorithms as the performance depends on many properties of datasets. But still it can be useful to identify some trends. One considers some aspects of data and tries to find some patterns in their relative performances.
 
In this work, we observe that for sparse datasets, L21 gave the best results generally with less number of features while on the other hand, while SC method gave the highest BSR with more number of features. These results can be justified by the fact that these datasets consist of many redundant features and the nature of SC method removes these redundant features. With the non-sparse data sets, LL method gave the highest BSR over all the presented methods.

Another observation was that out of EN and L21 which both use a combination of L1 and L2 norms, in most cases, EN gave higher BSR than L21 algorithm indicating that using both L1 and L2 norms with regularization part only is more productive than using it with both loss function and regularization.

On the parameter of correlation in the data, as expected, L1 doesn't do well for datasets with high values of correlation. This is on the expected lines. SC which is method that is based on the correlation does in fact give the best results as we would expect. But among the methods that we are comparing EN works the best, followed by L21. On datasets with low values of correlation, LL works the best in two out of the three datasets and L1 is the next best. This is again explained by the nature of L1 that it works best on less correlated datasets. Again though, EN is the best performer in one of the three least correlated datasets and the second best in one. This again indicates the `versatility' of EN for both highly correlated and less correlated data.  

On the parameter of sparsity, we observe that for dense datasets, LL, L1-SVM and EN gave the best results, while EN and LL gave the best results for sparse datasets. On the parameter of execution time, EN stands out as exceptionally better than other algorithms.

\bibliographystyle{plain}
\bibliography{Paper_10pg}

\end{document}